\documentclass[letterpaper, 10 pt, conference]{ieeeconf}  

\IEEEoverridecommandlockouts                              

\usepackage{graphics} 
\usepackage{epsfig} 
\usepackage{mathptmx} 
\usepackage{times} 
\usepackage{amsmath} 
\usepackage{amssymb,bm}  
\usepackage{graphicx}
\usepackage{color}
\usepackage{lipsum}

\usepackage[style=ieee,dashed=false]{biblatex}

\DeclareSourcemap{
  \maps{
    \map{
      \pertype{article}
      \step[fieldset=language, null]
      \step[fieldset=url, null]
      \step[fieldset=doi, null]
      \step[fieldset=issn, null]
      \step[fieldset=isbn, null]
      \step[fieldset=note, null]
      \step[fieldset=editor, null]
      \step[fieldset=urldate, null]
      \step[fieldset=file, null]
    }
  }
}
\DeclareSourcemap{
  \maps{
    \map{
      \pertype{inproceedings}
      \step[fieldset=language, null]
      \step[fieldset=url, null]
      \step[fieldset=doi, null]
      \step[fieldset=issn, null]
      \step[fieldset=isbn, null]
      \step[fieldset=note, null]
      \step[fieldset=editor, null]
      \step[fieldset=urldate, null]
      \step[fieldset=file, null]
    }
  }
}
\DeclareSourcemap{
  \maps{
    \map{
      \pertype{incollection}
      \step[fieldset=language, null]
      \step[fieldset=url, null]
      \step[fieldset=doi, null]
      \step[fieldset=issn, null]
      \step[fieldset=isbn, null]
      \step[fieldset=note, null]
      \step[fieldset=editor, null]
      \step[fieldset=urldate, null]
      \step[fieldset=file, null]
    }
  }
}

\title{\LARGE \bf
Banking Turn of High-DOF Dynamic Morphing Wing Flight by Shifting Structure Response Using Optimization
}

\author{Bibek Gupta$^{1}$, Yogi Shah$^{1}$, Taoran Liu$^{1}$, Eric Sihite$^{2}$ and Alireza Ramezani$^{1*}$
\thanks{$^{1}$Authors are with the Silicon Synapse Labs, Department of
Electrical and Computer Engineering, Northeastern University, Boston,
USA. Emails: gupta.bi, a.ramezani@northeastern.edu}%
\thanks{$^{2}$Author is with the Department of Aerospace Engineering,
California Institute of Technology, Pasadena, USA. Email:
esihite@caltech.edu}%
\thanks{*Corresponding author.}
}

\begin{document}

\maketitle
\thispagestyle{empty}
\pagestyle{empty}

\begin{abstract}
The 3D flight control of a flapping wing robot is a very challenging problem. The robot stabilizes and controls its pose through the aerodynamic forces acting on the wing membrane which has complex dynamics and it is difficult to develop a control method to interact with such a complex system. Bats, in particular, are capable of performing highly agile aerial maneuvers such as tight banking and bounding flight solely using their highly flexible wings. In this work, we develop a control method for a bio-inspired bat robot, the Aerobat, using small low-powered actuators to manipulate the flapping gait and the resulting aerodynamic forces. We implemented a controller based on collocation approach to track a desired roll and perform a banking maneuver to be used in a trajectory tracking controller. This controller is implemented in a simulation to show its performance and feasibility.

\end{abstract}

\section{Introduction}

{ 

Bats possess highly dynamic morphing wings which are known to be extremely high-dimensional with many active and passive modes \cite{riskin_quantifying_2008}. They can take advantage of their flexible wings to dynamically morph the shape of their wings to perform highly acrobatic maneuvers such as tight banking and turning maneuvers \cite{riskin_upstroke_2012,iriarte-diaz_whole-body_2011}. Copying bat dynamic morphing wing can bring fresh perspectives to micro aerial vehicle (MAV) design but is also extremely challenging \cite{ramezani_biomimetic_2017,ramezani_aerobat_2022,sihite_bang-bang_2022}.


The Northeastern University's \textit{Aerobat} platform, shown in Fig.~\ref{fig:aerobat-fulldyn}, is a bio-inspired bat robot that is developed to test dynamic morphing wing flight in realistic flight scenarios comparable to bat flights \cite{sihite_actuation_2023, dhole_hovering_2023, sihite_computational_2020}. This robot features a dynamically morphing wing where the wing can fold and expand during upstroke and downstroke, respectively, which improves flight efficiency. The robot armwing is composed of the linkages we called the ``computational structure'' which drives the robot's flapping gait. Some of the linkages in the computational structure are more sensitive where a small change in linkage length can greatly change the resulting flapping gait. In the overarching attempts to achieve high-dimensional actuation in flapping-wing MAVs, we utilized several low-powered actuators to manipulate the flapping gait of the robot, which in turn manipulates the generated aerodynamic forces acting on the wing \cite{sihite_actuation_2023, sihite_integrated_2021}. Using this method, we aim to implement an active closed-loop controller for flight trajectory tracking. The flight dynamic model of Aerobat \cite{sihite_unsteady_2022} and new contributions in this work are explained in more detail in Section \ref{sec:model}.


The framework we propose in this work has allowed the fast activation and regulation of many actuated degrees of freedom (DOF). We limit the scope of this work to simulation results to show that this framework is feasible to be used in a closed-loop trajectory-tracking controller. The implementation and testing of this gait regulation will be done in future work. This paper is outlined as follows: we begin with the background, motivations, and new contributions from our Aerobat platform and this work. Then, we followed with actuation methods and 3D trajectory tracking controller design based on collocation approach, their implementation in simulations, results discussion, and finalized with the concluding remarks.

}

\section{Flight Dynamics Model and Main Contributions}
\label{sec:model}

\subsection{Flight Dynamics Modelling}

\begin{figure}
    \centering
    \includegraphics[width=0.9\linewidth]{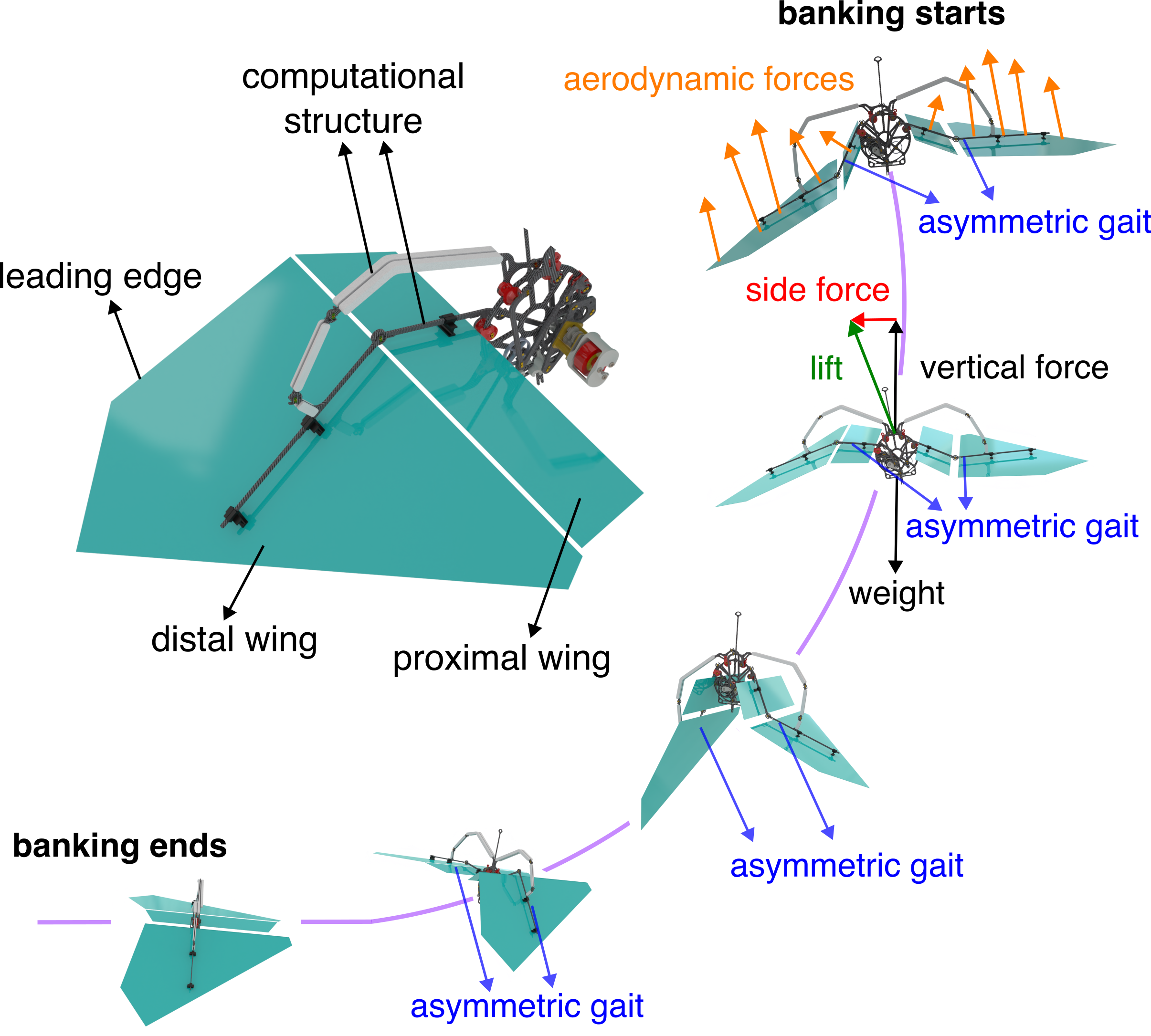}
    \caption{Cartoon idealization of Aerobat performing sharp banking turns.}
    \label{fig:aerobat-fulldyn}
\end{figure}

The flight dynamics of the Aerobat depicted in Fig.~\ref{fig:aerobat-fulldyn} can be elucidated through a cascade nonlinear system \cite{sihite_unsteady_2022} 
of the following form:
\begin{equation}
\begin{aligned}
\Sigma_{Full}&:\left\{
\begin{aligned}
    \dot x &= f(x) + g_1(x) u + g_2(x) y_2 \\
    y_1 &= h_1(x)\\
\end{aligned}
\right.\\
\Sigma_{Aero}&:\left\{
\begin{aligned}
    \dot \xi &= A_\xi (t) \xi + B_\xi (t) y_1 \\
    y_2 &= C_\xi (t)\xi + D_\xi (t) y_1
\end{aligned}
\right.
\end{aligned}
\label{eq:ss-rep-fulldyn}
\end{equation}
\noindent Here, $t$, $x$, and $\xi$ represent time, the state vector, and hidden aerodynamic variables, respectively. In Eq.~\ref{eq:ss-rep-fulldyn}, the state vector $x$ encompasses the position and velocity of both active $q_a$ and passive $q_p$ coordinates. The nonlinear terms denoted by $f(x)$ and $g_1(x)$ are derived from Lagrange equations and encompass inertial, Coriolis, and gravity effects \cite{sihite_unsteady_2022}. 
The state-dependent matrices $g_{1}(x)$ and $g_{2}(x)$ map the joint actions $u(x)$ and external force $y_2(x)$ to the state velocity vector $\dot{x}$.

The output $y_2$ represents the aerodynamic force output, providing information about the instantaneous external forces acting on Aerobat's wings during flapping wing flight. The governing dynamics are expressed in state-space form, characterized by matrices $A_\xi$, $B_\xi$, $C_\xi$, and $D_\xi$ \cite{sihite_unsteady_2022}.
These terms are derived from the Wagner indicial model and Prandtl lifting line theory, foundational concepts in fluid dynamics \cite{sihite_unsteady_2022}. The efficiency of the indicial model lies in two merits:
\begin{itemize}
    \item First, indicial models can predict unsteady aerodynamic forces-moments efficiently, which is an important feat for optimization-based control of flying systems.  
    \item Second, indicial models can predict wake structures based on horseshoe vortex shedding. This capability facilitates the description of locomotion gaits using wake structures. Such gaits are extensively utilized in biology to characterize bat aerial locomotion \cite{hubel_wake_2010,parslew_theoretical_2013}. 
\end{itemize}

Next, we briefly discuss the significance of the cascade model described by Eq.~\ref{eq:ss-rep-fulldyn} from the perspective of actuation in small robots such as Aerobat, which cannot accommodate a large number of power actuators.

\subsection{Paper's Main Contributions}

\begin{figure}
    \centering
    \includegraphics[width=1.0\linewidth]{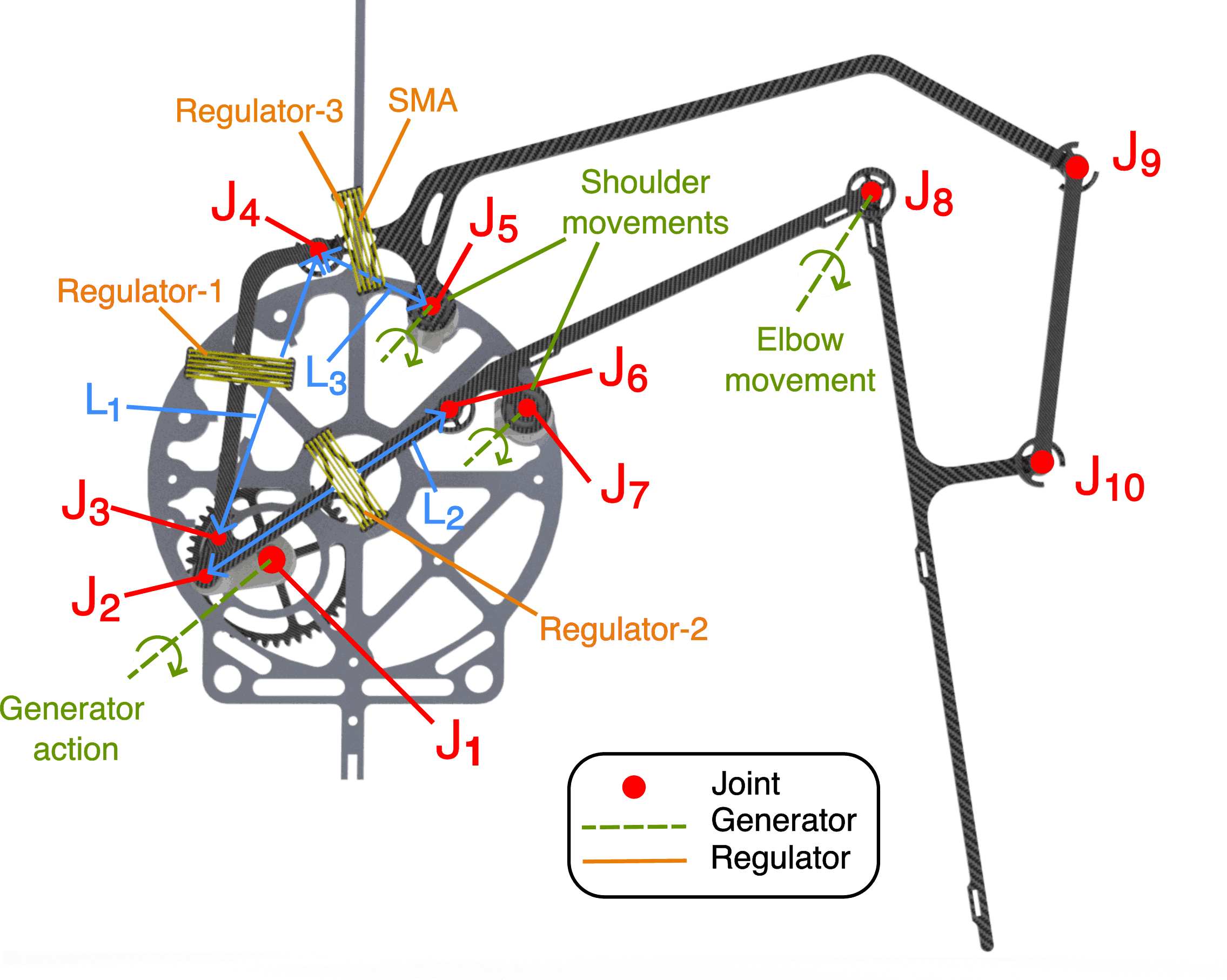}
    \caption{Shows the computational structure used in Aerobat and the proposed low-power actuators' locations within the structure (regulator). Also shows the relevant joints (shoulder and elbow) that generate the resulting flapping gait (generator).
    }
    \label{fig:ks}
\end{figure}

The main goal of this study is to execute intricate 3D flight maneuvers using Aerobat, despite its very limited actuation resources. Achieving such maneuvers necessitates the robot's adept manipulation of its fluidic surroundings (wing plunge motion, elbow flexion-extension movements, feathering, sweeping, etc.), primarily accomplished through gait control. In Eq.~\ref{eq:ss-rep-fulldyn}, the key functions of control inputs include (1) generating gaits, involving swift movements of cluster joints, and (2) regulating these movements, ensuring precise adjustments within the cluster joints. Dynamic morphing in Aerobat imposes stringent demands on the input vector, encompassing requirements such as power density and the challenge of dealing with small space and high-dimensional joint space. 

The cascade model given by Eq.~\ref{eq:ss-rep-fulldyn} implicitly integrates the responsibility of input $u$ within mechanical intelligence and controls (for a brief discussion see Section~\ref{sec:gen-reg}). This means that by incorporating the holonomic constraint $y_1=h_1(x)$, we create a mathematical framework to systematically split the contributions of input $u=\left[u_{Reg}^\top,u_{Gen}^\top\right]^\top$ (see \cite{ramezani_biomimetic_2017,sihite_actuation_2023}).

While there hasn't been a clear strategy in the literature for systematically splitting actuator roles for motion control in small robots unable to accommodate large actuators, motion control strategies, assuming a uniform role for all actuators (i.e., all entries of the input vector $u_i$ recruited in the same manner), have been common in larger systems like manipulators and legged systems, which face fewer design constraints. This study stands out as it delves into motion control for complex robots with articulated bodies, such as Aerobat, by assigning different roles to actuators. This perspective can reduce the dependency on large actuators, which are challenging to accommodate in small systems, while allowing for motion control characteristics such as precision and speed observed in larger systems. 

\section{Generators and Regulators}
\label{sec:gen-reg}

As illustrated in Fig.~\ref{fig:ks}, Aerobat's wings feature numerous moving joints. The central question addressed here is how to mobilize these joints effectively. Two approaches exist for generating and manipulating these joint motions.

\begin{itemize}
    \item Firstly, we can consider separate, isolated actuated joints. While effective for proximal joints, this approach may impose significant inertial loads and potential failure at distal points \cite{sihite_enforcing_2020}.
    \item Alternatively, we can employ a constrained mechanical structure, represented by $y_1=h_1(x)$, which interconnects joint motions for dynamic motion control. For manipulation of joint motion, the response of the constrained mechanical structure can be adjusted using small, low-power actuators embedded within it.
\end{itemize}

We propose the utilization of the second approach. The design of Aerobat, as depicted in Fig.~\ref{fig:ks}, is based on this second option. For more details about Aerobat's hardware, readers are referred to \cite{sihite_unsteady_2022}. 


In this study, we aim to showcase how adjusting the positions of regulators with a fast and computationally efficient controller can facilitate the manipulation of correlated joint motions $y_1=h_1(x)$ in Aerobat, consequently enabling force tracking. We tackle this tracking challenge by employing a collocation optimization method that enables rapid approximation of the computational structure dynamics. It's imperative to highlight the significance of speed and efficiency in control calculations, especially in dynamic morphing wing flight scenarios where body joint motions occur rapidly, and there is limited space to accommodate powerful computers.

\section{Efficient 3D Path Tracking Controls Based on Collocation Approach}

To solve this flight control problem, i.e., Aerobat's posture $y_1$ is recruited to regulate fluid-structure forces-moments $y_2$ to track a 3D path, we consider the following cost function given by
%
\begin{equation}
    J = \sum_i^N (z_{i} - z_{ref,i})^\top \, C \, (z_{i} - z_{ref,i}),
    \quad
    z = [\theta_r, \theta_p, \omega^\top]^\top,
    \label{eq:cost}
\end{equation}
%
{
where $\theta_{r}$ and $\theta_{p}$ represent the robot's roll and pitch angles relative to the inertial frame, respectively, while $\omega$ represents the robot's angular velocities, $z$ is the optimization state, $z_{ref}$ is the state reference for $z$, and $C$ is a diagonal cost weighting matrix. The cost function $J$ is governed by a system of $n$ nonlinear equations representing the computational structure dynamics driven by low-power actuators, as depicted in Fig.~\ref{fig:ks}.
}

\color{black}
To further elucidate, following the principle of virtual work \cite{landau1982mechanics}, the response from the computational structure can be determined by
\begin{equation}
\begin{aligned}
    \begin{bmatrix}
    \dot y_{1,1}\\
    \ddot y_{1,1}\\
    \vdots\\
    \dot y_{1,n}\\
    \ddot y_{1,n}\\
    \end{bmatrix}=&
    \begin{bmatrix} 
    a_{1,1} & a_{1,2} & \dots \\
    \vdots & \ddots & \\
    a_{2n,1} &        & a_{2n,2n} 
    \end{bmatrix}
    \begin{bmatrix}
    y_{1,1}\\
    \dot y_{1,1}\\
    \vdots\\
    y_{1,n}\\
    \dot y_{1,n}\\
    \end{bmatrix}
    +\\
    & \hspace{1cm} \begin{bmatrix} 
    b_{1,1} & b_{1,2} & \dots \\
    \vdots & \ddots & \\
    b_{2n,1} &        & b_{2n,m} 
    \end{bmatrix}
    \begin{bmatrix}
    \omega_1\\
    \vdots\\
    \omega_m\\
    \end{bmatrix}
\end{aligned}
    \label{eq:mimic_dynamics}
\end{equation}
where $y_{1,j}, ~j=1, \ldots, n$ denotes the movement from each element of the computational structure, $a_{j,k}$ and $b_{j,k}$ are determined by the physical properties, and $\omega_j,~j=1, \ldots, m$ is the regulator's input. 

By inspecting Eq.~\ref{eq:mimic_dynamics}, it can be observed that the contribution of the input term $u$ based on mode generation and regulation can be separately considered through the design of $a_{j,k}$ (i.e., structure configuration and material properties) and $b_{j,k}$ (regulator or low-power actuator placement), as discussed in \cite{sihite_actuation_2023}.

We perform temporal (i.e., $t_i,~i=1, \ldots, n, \quad 0 \leq t_i \leq t_f$) discretization of Eq.~\ref{eq:mimic_dynamics} to obtain the following system of equations
\begin{equation}
\dot{Y}_{i}(t_i)=A_{i}Y_i(t_i) + B_{i}\Omega_i(t_i), \quad i=1, \ldots, n, \quad 0 \leq t_i \leq t_f
\label{eq:disc-model}    
\end{equation}
where $Y_i =\left[y_{1,1}^\top,\dots,y_{1,n}^\top\right]^\top$ embodies $n$ spatial values of the computational structure response at i-th discrete time (i.e., posture at time $t_i$). And, $\Omega_i=\left[\omega_1,\dots,\omega_m\right]$ embodies $m$ regulators actions at i-th discrete time. $A_i$ and $B_i$ are the matrices shown in Eq.~\ref{eq:mimic_dynamics} with their entries. 

We stack all of the postures and low-power inputs from the regulators from each i-th sample time, i.e., $Y_i$ and $\Omega_i$, in the vectors $Y = \left[Y^\top_1(t_1), \ldots, Y_n^\top(t_n)\right]^\top$ and $\Omega = \left[\Omega^\top_1(t_1), \ldots, \Omega^\top_m(t_n)\right]^\top$. 

We consider 2$n$ boundary conditions at the boundaries of $n$ structure elements (2 equations at each boundary) to enforce the continuity of the computational structure, given by
\begin{equation}
    r_i\left(Y(0), Y\left(t_f\right), t_f\right)=0, \quad i=1, \ldots, 2 n
\end{equation}
Since we have $m$ regulators, we consider $m$ inequality constraints given by
\begin{equation}
    g_i(Y(t_i), \Omega(t_i), t_i) \geq 0, \quad i=1, \ldots, m, \quad 0 \leq t_i \leq t_f
\end{equation}
to limit the actuation stroke from the low-power actuators.  

To approximate nonlinear dynamics from the computational structure, we employ a method based on polynomial interpolations. This method extremely simplifies the computation efforts.  

\subsection{Polynomial Approximation of the Discrete States and Regulator Inputs}

Consider the $n$ time intervals during a gait cycle of the dynamic morphing system, as defined previously and given by 
\begin{equation}
0=t_1<t_2<\ldots<t_n=t_f
\end{equation}
We stack the states and regulator inputs $Y = \left[Y^\top_1(t_1), \ldots, Y_n^\top(t_n)\right]^\top$ and $\Omega = \left[\Omega^\top_1(t_1), \ldots, \Omega^\top_m(t_n)\right]^\top$ from the computational structure at these discrete times into a single vector denoted by $\mathcal{Y}$ and form a decision parameter vector that the optimizer finds at once. Additionally, we append the final discrete time $t_f$ as the last entry of $\mathcal{Y}$ so that gaitcycle time too is determined by the optimizer.
\begin{equation}
\mathcal{Y}=\left[Y^\top_1(t_1), \ldots, Y_n^\top(t_n), \Omega^\top_1(t_1), \ldots, \Omega^\top_m(t_n), t_f\right]^\top
\end{equation}
We approximate the regulator's action at time $t_i \leq t<t_{i+1}$ as the linear interpolation function $\tilde\Omega(t)$ between $\Omega_i(t_i)$ and $\Omega_{i+1}(t_{i+1})$ given by 
\begin{equation}
\tilde\Omega(t)=\Omega_i\left(t_i\right)+\frac{t-t_i}{t_{i+1}-t_i}\left(\Omega_{i+1}\left(t_{i+1}\right)-\Omega_i\left(t_i\right)\right)
\end{equation}
We interpolate the computational structure states $Y_i(t_i)$ and $Y_{i+1}(t_{i+1})$ as well. However, we use a nonlinear cubic interpolation, which is continuously differentiable with $\dot{\tilde Y}(s)=\bm F(Y(s), \Omega(s), s)$ at $s=t_i$ and $s=t_{i+1}$, where $\bm F$ is formed from the dynamics defined in Eq.~\eqref{eq:ss-rep-fulldyn}. 

To obtain $\tilde Y(t)$, we formulate the following system of equations:
\begin{equation}
    \begin{aligned}
\tilde Y(t) &=\sum_{k=0}^3 c_k^j\left(\frac{t-t_j}{h_j}\right)^k, \quad t_j \leq t<t_{j+1}, \\
c_0^j &=Y\left(t_j\right), \\
c_1^j &=h_j \bm F_j, \\
c_2^j &=-3 Y\left(t_j\right)-2 h_j \bm F_j+3 Y\left(t_{j+1}\right)-h_j \bm F_{j+1}, \\
c_3^j &=2 Y\left(t_j\right)+h_j \bm F_j-2 Y\left(t_{j+1}\right)+h_j \bm F_{j+1}, \\
\text { where } \bm F_j &:=\bm F\left(Y\left(t_j\right), \Omega\left(t_j\right), t_j\right), \quad h_j:=t_{j+1}-t_j .
\end{aligned}
\label{eq:cubic-lobatto}
\end{equation}
The interpolation function $\tilde Y$ utilized for $Y$ needs to fulfill the computational structure's derivative requirements at discrete points and at the midpoint of sample times. By examining Eq.~\ref{eq:cubic-lobatto}, it is evident that the derivative terms at the boundaries $t_{i}$ and $t_{i+1}$ are satisfied. Hence, the only remaining constraints in the nonlinear programming problem are the collocation constraints at the midpoint of $t_i-t_{i+1}$ time intervals, the inequality constraints at $t_i$, and the constraints at $t_1$ and $t_f$, all of which are included in the optimization process.

Given that the computational structure is spatially discrete and incurs significant costs associated with its curse of dimensionality, this collocation scheme reduces the number of parameters for interpolation polynomials, thereby enhancing computational performance. We address this optimization problem using MATLAB's fmincon function.  

{
\section{Results}

\begin{figure}[t]
  \centering
  \includegraphics[width=0.8\columnwidth]{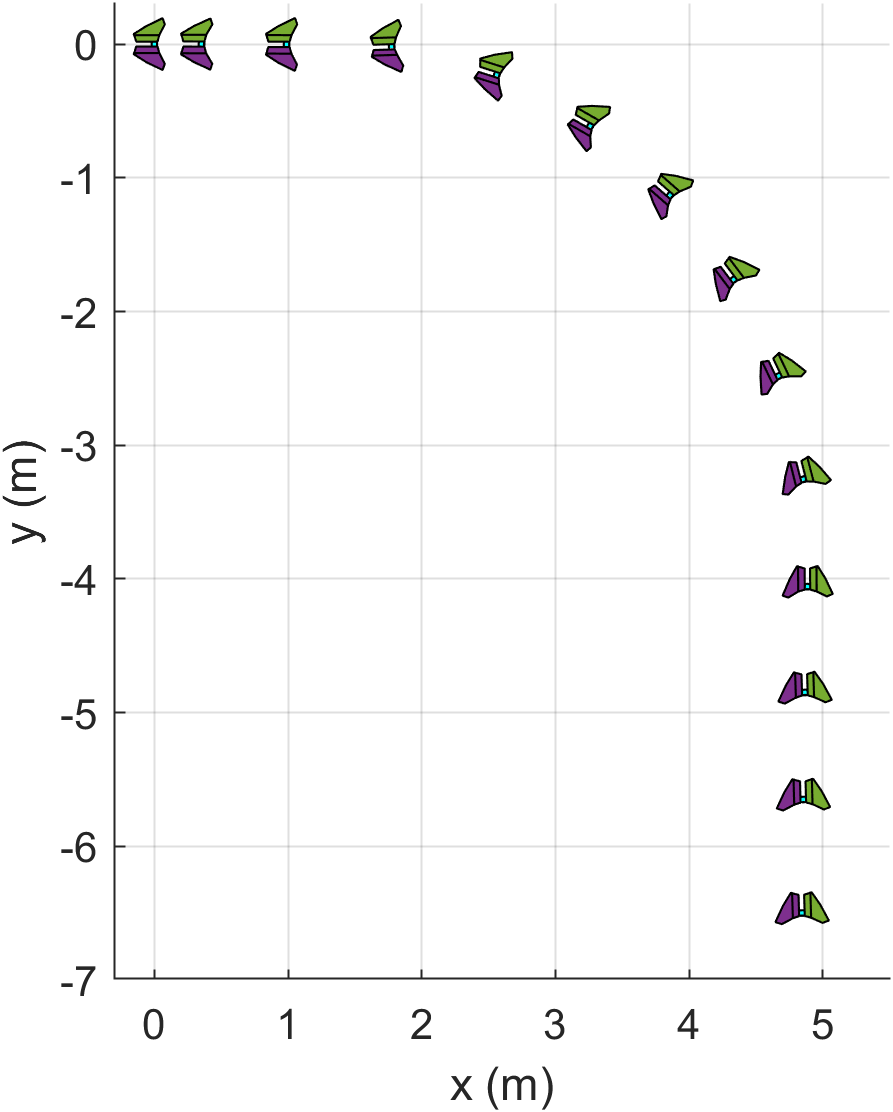}
  \caption{Top view snapshots of the simulation showing the trajectory of the banking turn.}
  \label{fig:banking}
\end{figure}

\begin{figure*}[t]
  \centering
  \includegraphics[width=0.9\linewidth]{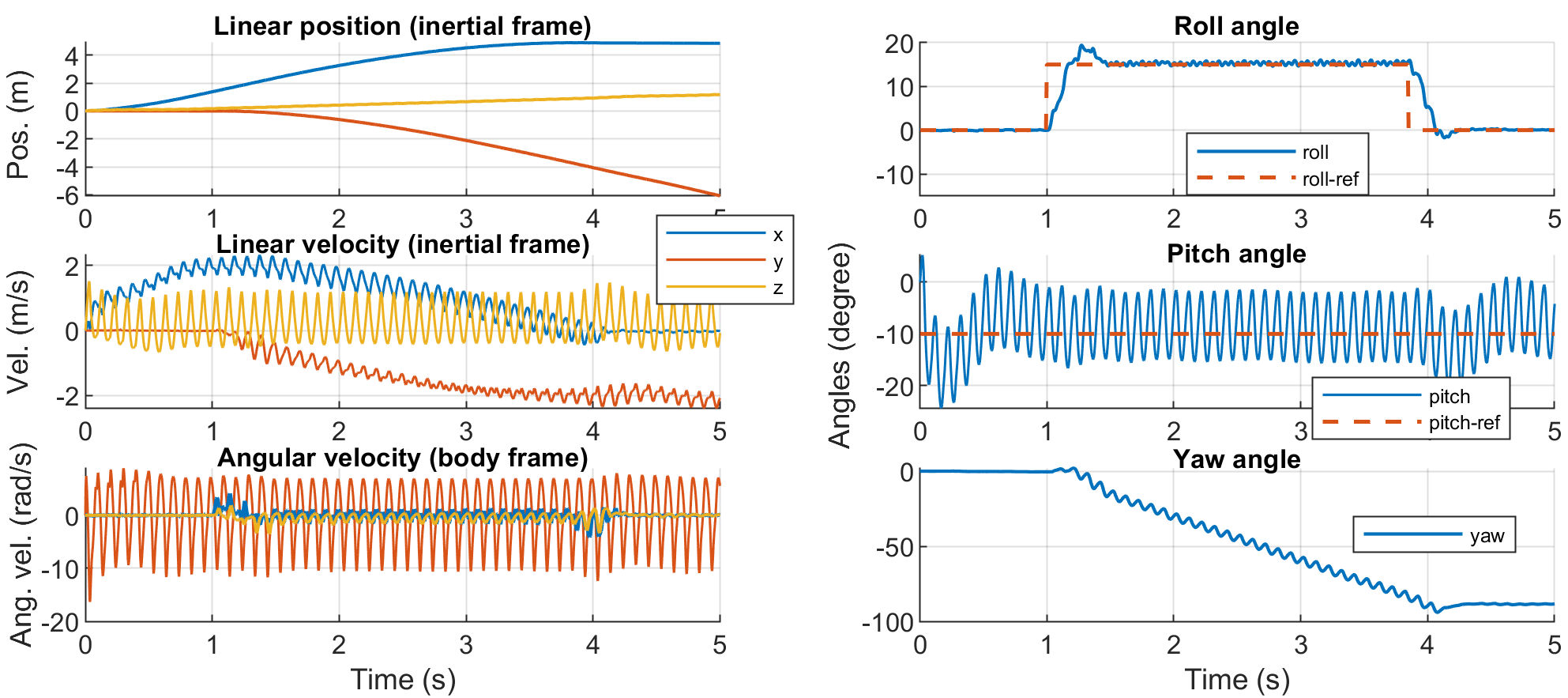}
  \caption{Plots of the robot's position and orientation states versus time in the simulation. The robot has successfully tracked the desired roll to perform the banking maneuver and generated a steady change in the robot's heading.}
  \label{fig:states}
\end{figure*}

\begin{figure*}[t]
  \centering
  \includegraphics[width=0.9\linewidth]{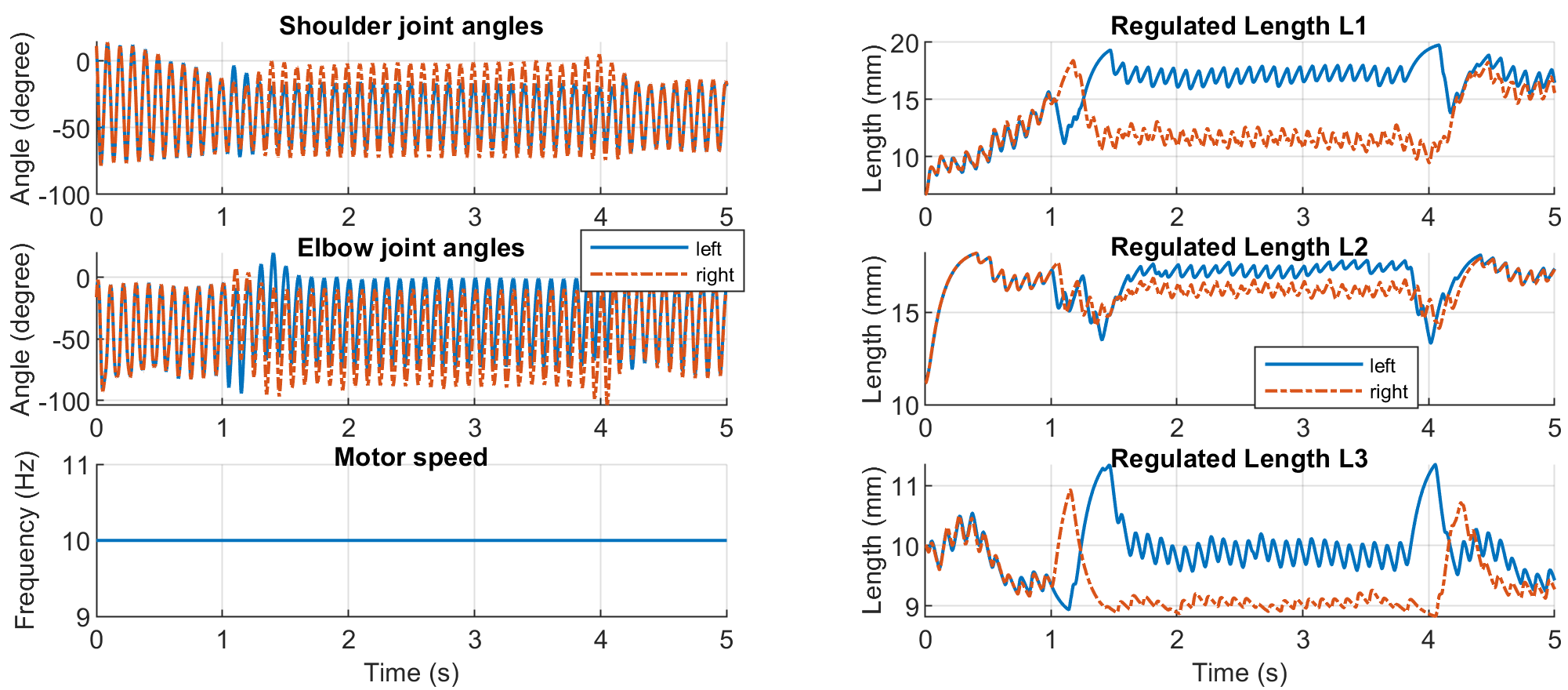}
  \caption{Plots of the robot's generator trajectories (shoulder and elbow joint angles) and regulator lengths versus time in the simulation. The plots show how the asymmetric flapping gait and regulator actions between the left and right wings that allow the robot to perform the banking maneuver.}
  \label{fig:fdc_states}
\end{figure*}

\begin{figure*}[t]
    \centering
    \includegraphics[width=0.9\linewidth]{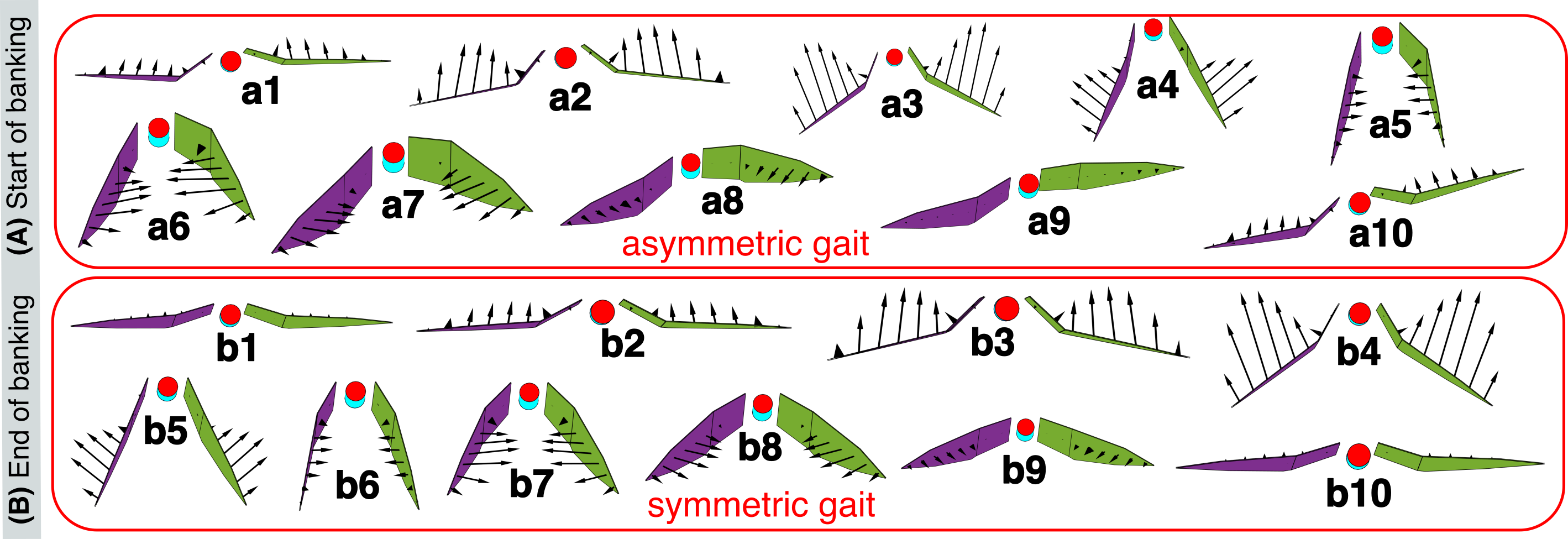}
    \caption{Simulation snapshots of the frontal view of the robot and the generated aerodynamic forces are shown applied on the wings. The gait is asymmetric during the banking maneuver, causing the robot to bank and turn. 
    }
    \label{fig:gaits}
\end{figure*}

\begin{figure*}[t]
    \centering
    \includegraphics[width=\linewidth,height=8cm]{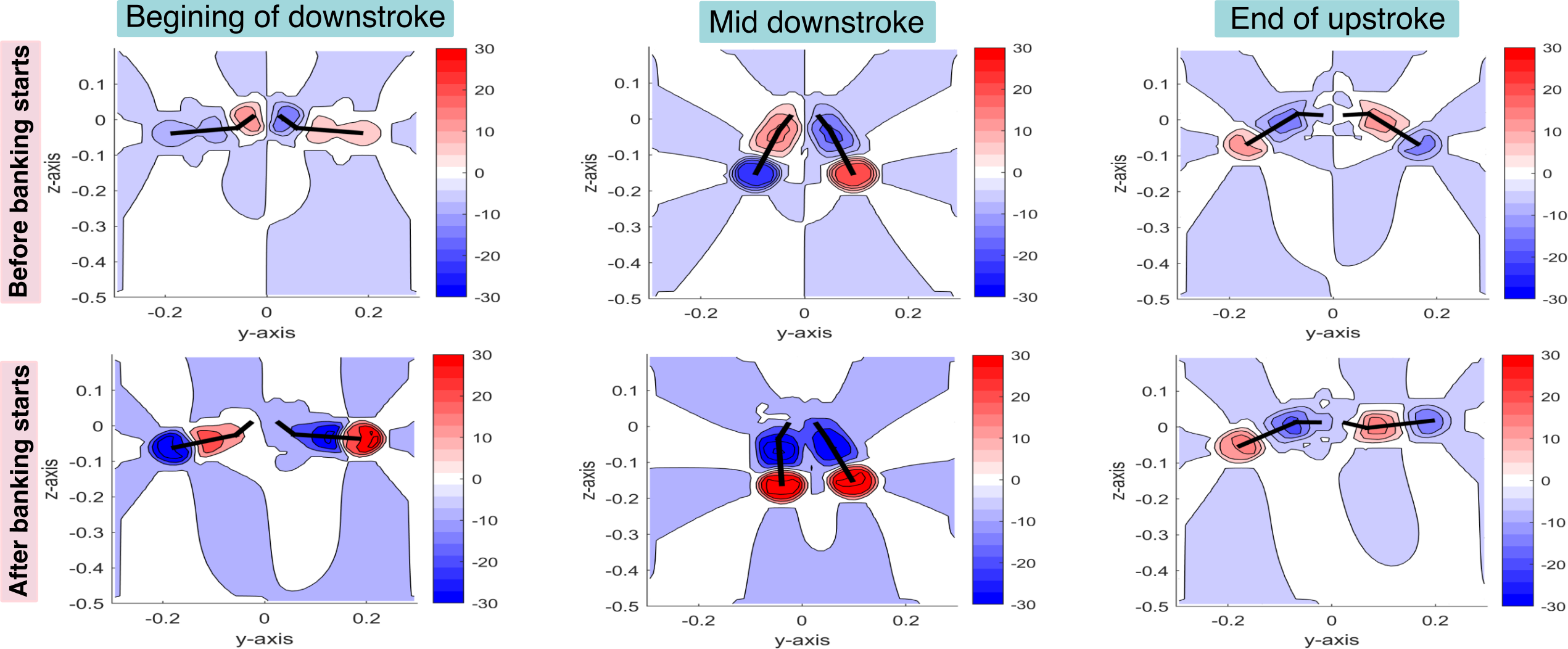}
    \caption{
    Phase map plots showing the magnitude of the vortices induced by the flapping gait in rad/s, where the vorticity levels near the wing tip show higher vorticity during downstrokes and lower during upstrokes. Black lines represent Aerobat's wing in frontal view. 
    }
    \label{fig:vorticity_plots}
\end{figure*}

The modeling and simulation of the non-linear dynamical system, as described by the equations presented in Eq.~\ref{eq:ss-rep-fulldyn}, were conducted using MATLAB. The system of equations was solved numerically employing the fourth-order Runge-Kutta method, a well-regarded technique for its accuracy and computational efficiency. A collocation-based optimization controller, incorporating the complete nonlinear dynamic equation of motion within its predictive framework, was utilized. To enhance computational efficiency, a 5-step prediction horizon was adopted. The simulation was conducted with a time step of 0.0001 seconds, while the collocation controller's update time step was set at 0.005 seconds (200 Hz). 

In the simulation, the motor—functioning as a generator—produced flapping gaits at a constant rate of 10 Hz. The regulators adjusted the lengths of links 1, 2, and 3, modifying the flapping gaits to achieve the desired roll and pitch references. Specifically, the pitch reference was maintained at a constant negative 10 degrees throughout the simulation. Meanwhile, the roll reference was initially set to 0 degrees for the first second, adjusted to 15 degrees for the subsequent 2.8 seconds to facilitate the banking turn, and then reverted to 0 degrees for the remainder of the simulation time.

The results elucidated in Fig.~\ref{fig:states} and Fig.~\ref{fig:fdc_states} offer a detailed examination of the system's dynamic states and the regulation of link lengths, alongside the precision of Euler angle tracking. Figure~\ref{fig:banking}, in particular, showcases the collocation controller's proficiency in adhering to predefined roll and pitch angles. It is observed that the robot's pitch fluctuates around a constant -10 degrees with an amplitude in the range of 8 to 10 degrees. This consistent oscillation is caused by the wing's flapping motion.

Moreover, the roll reference tracking is maintained with a high degree of accuracy, facilitating a significant yaw angle alteration that results in a smooth banking turn, as depicted in fig.~\ref{fig:banking}. Over the full trajectory span of 5 seconds, the robot successfully ascended to a height of 1.2 meters and traversed a distance of 4.8 meters in the x-direction and 6 meters in the y-direction. This demonstrates the system's integrated control strategy's capability to achieve complex maneuvers, such as coordinated turns, with remarkable spatial efficiency.

The success of the aerial maneuvers can be attributed to the precise regulation of link lengths within the robot's wing structure, which is essential for modifying the flapping patterns. Figure
~\ref{fig:fdc_states} clearly illustrates the variations in length between the robot's right and left wings. These adjustments result in changes to the shoulder and elbow angles across the wings, consequently leading to asymmetric gaits. 

Such asymmetry is crucial for the robot's ability to execute complex maneuvers, like banking turns, by allowing differential aerodynamic forces to be generated on either side of the robot. Figure~\ref{fig:gaits} vividly captures the robot engaging in these asymmetric flapping gaits during such maneuvers. Conversely, when the trajectory necessitates straight-line motion, the robot reverts to symmetrical flapping gaits, demonstrating its adaptable wing movement for effective navigation.

Figure~\ref{fig:vorticity_plots} presents the simulated vorticity in the vicinity of the wingtip, capturing the dynamic interaction between the Aerobat and the surrounding air throughout distinct stages of a flapping cycle. Notably, the vorticity is significantly higher during the downstroke, indicative of stronger aerodynamic forces at play, while it diminishes when the wings fold during the upstroke, reflecting a decrease in aerodynamic activity. Before banking, the vorticity distribution is relatively symmetrical, signifying a steady flight. However, as banking commences, an asymmetrical vorticity pattern emerges, with marked differences between the two sides of the Aerobat. This asymmetry is crucial for the initiation and maintenance of the banking maneuver, altering the lift and drag forces on the wings to achieve the desired roll angle for the turn.

}


{
\color{blue}
}

\section{Concluding Remarks}

In this paper, we present our control framework to perform a closed-loop trajectory tracking of a flapping wing robot, the Aerobat. The robot utilizes three regulators on each wing and a single motor to drive and generate the flapping gait for stabilizing the robot's orientation in the air. The application of the collocation control framework was shown in the simulation where we successfully tracked the desired body roll and pitch angles mid-flight to perform the bank turning maneuver. This approach can further be extended to track a predefined flight path for a more practical application such as exploration and reconnaissance. For future work, we will develop and implement the regulator into our existing Aerobat platform and eventually perform untethered experiments to show the feasibility of this controller framework. 


\printbibliography

@inproceedings{sihite_actuation_2023,
	address = {Singapore},
	title = {Actuation and {Flight} {Control} of {High}-{DOF} {Dynamic} {Morphing} {Wing} {Flight} by {Shifting} {Structure} {Response}},
	booktitle = {Conference on {Decision} and {Control} ({CDC})},
	author = {Sihite, Eric and Salagame, Adarsh and Ghanem, Paul and Ramezani, Alireza},
	month = dec,
	year = {2023},
}

@misc{dhole_hovering_2023,
	title = {Hovering {Control} of {Flapping} {Wings} in {Tandem} with {Multi}-{Rotors}},
	url = {http://arxiv.org/abs/2308.00183},
	doi = {10.48550/arXiv.2308.00183},
	urldate = {2023-10-19},
	publisher = {arXiv},
	author = {Dhole, Aniket and Gupta, Bibek and Salagame, Adarsh and Niu, Xuejian and Xu, Yizhe and Venkatesh, Kaushik and Ghanem, Paul and Mandralis, Ioannis and Sihite, Eric and Ramezani, Alireza},
	month = jul,
	year = {2023},
	note = {arXiv:2308.00183 [cs, eess]},
}

@inproceedings{sihite_integrated_2021,
	title = {An {Integrated} {Mechanical} {Intelligence} and {Control} {Approach} {Towards} {Flight} {Control} of {Aerobat}},
	doi = {10.23919/ACC50511.2021.9482643},
	booktitle = {2021 {American} {Control} {Conference} ({ACC})},
	author = {Sihite, Eric and Darabi, Atefe and Dangol, Pravin and Lessieur, Andrew and Ramezani, Alireza},
	month = may,
	year = {2021},
	note = {ISSN: 2378-5861},
	pages = {84--91},
}

@inproceedings{sihite_unsteady_2022,
	title = {Unsteady aerodynamic modeling of {Aerobat} using lifting line theory and {Wagner}'s function},
	doi = {10.1109/IROS47612.2022.9982125},
	booktitle = {2022 {IEEE}/{RSJ} {International} {Conference} on {Intelligent} {Robots} and {Systems} ({IROS})},
	author = {Sihite, Eric and Ghanem, Paul and Salagame, Adarsh and Ramezani, Alireza},
	month = oct,
	year = {2022},
	note = {ISSN: 2153-0866},
	pages = {10493--10500},
}

@misc{sihite_bang-bang_2022,
	title = {Bang-{Bang} {Control} {Of} {A} {Tail}-less {Morphing} {Wing} {Flight}},
	url = {http://arxiv.org/abs/2205.06395},
	doi = {10.48550/arXiv.2205.06395},
	urldate = {2023-05-17},
	publisher = {arXiv},
	author = {Sihite, Eric and Hu, Xintao and Li, Bozhen and Salagame, Adarsh and Ghanem, Paul and Ramezani, Alireza},
	month = may,
	year = {2022},
	note = {arXiv:2205.06395 [cs, eess]},
}

@misc{ramezani_aerobat_2022,
	title = {Aerobat, {A} {Bioinspired} {Drone} to {Test} {High}-{DOF} {Actuation} and {Embodied} {Aerial} {Locomotion}},
	url = {http://arxiv.org/abs/2212.05361},
	doi = {10.48550/arXiv.2212.05361},
	urldate = {2023-05-17},
	publisher = {arXiv},
	author = {Ramezani, Alireza and Sihite, Eric},
	month = dec,
	year = {2022},
	note = {arXiv:2212.05361 [cs, eess]},
}

@article{hubel_wake_2010,
	title = {Wake structure and wing kinematics: the flight of the lesser dog-faced fruit bat, {Cynopterus} brachyotis},
	volume = {213},
	issn = {0022-0949},
	shorttitle = {Wake structure and wing kinematics},
	url = {https://doi.org/10.1242/jeb.043257},
	doi = {10.1242/jeb.043257},
	number = {20},
	urldate = {2021-08-04},
	journal = {Journal of Experimental Biology},
	author = {Hubel, Tatjana Y. and Riskin, Daniel K. and Swartz, Sharon M. and Breuer, Kenneth S.},
	month = oct,
	year = {2010},
	pages = {3427--3440},
}

@article{parslew_theoretical_2013,
	title = {Theoretical modelling of wakes from retractable flapping wings in forward flight},
	volume = {1},
	issn = {2167-8359},
	url = {https://peerj.com/articles/105},
	doi = {10.7717/peerj.105},
	language = {en},
	urldate = {2021-07-31},
	journal = {PeerJ},
	author = {Parslew, Ben and Crowther, William J.},
	month = jul,
	year = {2013},
	note = {Publisher: PeerJ Inc.},
	pages = {e105},
}

@article{riskin_quantifying_2008,
	title = {Quantifying the complexity of bat wing kinematics},
	volume = {254},
	issn = {0022-5193},
	url = {https://www.sciencedirect.com/science/article/pii/S002251930800324X},
	doi = {10.1016/j.jtbi.2008.06.011},
	language = {en},
	number = {3},
	urldate = {2021-07-04},
	journal = {Journal of Theoretical Biology},
	author = {Riskin, Daniel K. and Willis, David J. and Iriarte-Díaz, José and Hedrick, Tyson L. and Kostandov, Mykhaylo and Chen, Jian and Laidlaw, David H. and Breuer, Kenneth S. and Swartz, Sharon M.},
	month = oct,
	year = {2008},
	pages = {604--615},
}

@inproceedings{sihite_enforcing_2020,
	address = {Jeju Island, Korea (South)},
	title = {Enforcing nonholonomic constraints in {Aerobat}, a roosting flapping wing model},
	isbn = {978-1-72817-447-1},
	url = {https://ieeexplore.ieee.org/document/9304158/},
	doi = {10.1109/CDC42340.2020.9304158},
	language = {en},
	urldate = {2021-03-01},
	booktitle = {2020 59th {IEEE} {Conference} on {Decision} and {Control} ({CDC})},
	publisher = {IEEE},
	author = {Sihite, Eric and Ramezani, Alireza},
	month = dec,
	year = {2020},
	pages = {5321--5327},
}

@article{sihite_computational_2020,
	title = {Computational {Structure} {Design} of a {Bio}-{Inspired} {Armwing} {Mechanism}},
	volume = {5},
	issn = {2377-3766, 2377-3774},
	url = {https://ieeexplore.ieee.org/document/9143405/},
	doi = {10.1109/LRA.2020.3010217},
	language = {en},
	number = {4},
	urldate = {2021-03-01},
	journal = {IEEE Robotics and Automation Letters},
	author = {Sihite, Eric and Kelly, Peter and Ramezani, Alireza},
	month = oct,
	year = {2020},
	pages = {5929--5936},
}

@article{ramezani_biomimetic_2017,
	title = {A biomimetic robotic platform to study flight specializations of bats},
	volume = {2},
	issn = {2470-9476},
	url = {https://robotics.sciencemag.org/lookup/doi/10.1126/scirobotics.aal2505},
	doi = {10.1126/scirobotics.aal2505},
	language = {en},
	number = {3},
	urldate = {2021-03-01},
	journal = {Science Robotics},
	author = {Ramezani, Alireza and Chung, Soon-Jo and Hutchinson, Seth},
	month = feb,
	year = {2017},
	pages = {eaal2505},
}

@article{iriarte-diaz_whole-body_2011,
	title = {Whole-body kinematics of a fruit bat reveal the influence of wing inertia on body accelerations},
	volume = {214},
	language = {en},
	number = {9},
	urldate = {2019-09-04},
	journal = {Journal of Experimental Biology},
	author = {Iriarte-Diaz, J. and Riskin, D. K. and Willis, D. J. and Breuer, K. S. and Swartz, S. M.},
	month = may,
	year = {2011},
	pages = {1546--1553},
}

@article{riskin_upstroke_2012,
	title = {Upstroke wing flexion and the inertial cost of bat flight},
	volume = {279},
	language = {eng},
	number = {1740},
	journal = {Proceedings. Biological Sciences},
	author = {Riskin, Daniel K. and Bergou, Attila and Breuer, Kenneth S. and Swartz, Sharon M.},
	month = aug,
	year = {2012},
	pages = {2945--2950},
}

@book{landau1982mechanics,
  title={Mechanics: Volume 1},
  author={Landau, L.D. and Lifshitz, E.M.},
  number={v. 1},
  isbn={9780080503479},
  url={https://books.google.com/books?id=bE-9tUH2J2wC},
  year={1982},
  publisher={Elsevier Science}
}

\end{document}